\documentclass[11pt]{article}

\usepackage[preprint]{acl}

\usepackage{times}
\usepackage{latexsym}
\usepackage{amsmath}
\usepackage[inline]{enumitem}
\usepackage{mdframed}
\usepackage[dvipsnames]{xcolor}
\usepackage[commandnameprefix=always]{changes}
\usepackage[most]{tcolorbox}
\usepackage{booktabs}
\usepackage{array}
\usepackage{pdfpages}

\usepackage[T1]{fontenc}

\usepackage[utf8]{inputenc}

\usepackage{microtype}

\usepackage{inconsolata}

\usepackage{graphicx}

\usepackage{color-edits}
\addauthor[Emily]{es}{teal}
\addauthor[Dhruv]{da}{red}
\addauthor[Solon]{sib}{purple}
\addauthor[Hanna]{hmw}{green}

\newcommand{\spec}[1]{\texttt{#1-spec}}
\newcommand{\artifacts}[1]{\texttt{#1-artifacts}}

\title{AI-Assisted Systematization for Evaluating GenAI Systems}

\author{
 \textbf{Dhruv Agarwal}\textsuperscript{1}\thanks{~~Equal contribution.}\thanks{~~Correspondance: \href{mailto:da399@cornell.edu}{da399@cornell.edu}} \quad
 \textbf{Emily Sheng}\textsuperscript{2}\footnotemark[1]
\\
 \textbf{Chad Atalla}\textsuperscript{2} \quad
 \textbf{Jean Garcia-Gathright}\textsuperscript{2} \quad
 \textbf{Hussein Mozannar}\textsuperscript{2} \quad
 \textbf{Hannah Washington}\textsuperscript{2}
\\
 \textbf{Alexandra Chouldechova}\textsuperscript{2} \quad
 \textbf{Solon Barocas}\textsuperscript{2} \quad
 \textbf{Hanna Wallach}\textsuperscript{2}
\\
\\
 \textsuperscript{1}Cornell University \quad
 \textsuperscript{2}Microsoft Research
}

\begin{document}
\maketitle

\begin{abstract}
Evaluating generative AI (GenAI) systems is challenging because many targets of evaluation are broad, contested concepts, such as ``reasoning,'' ``fairness,'' or ``creativity.''
When these concepts are left underspecified, it becomes unclear what should be measured or how evaluation results should be interpreted. This problem reflects a missing step: \textit{systematization}, that is, moving from a broad background concept to an explicit, structured account of the concept in measurable terms. To help address the fact that systematization is cognitively demanding and resource-intensive, we investigate whether AI assistance can support this process. To enable AI-assisted systematization and assess its quality, we introduce a structured representation of a systematized concept, a \textit{concept spec}, and a validation worksheet. We then develop two AI-assisted systematizers: a direct, zero-shot approach and a multi-agent approach that more closely mirrors manual systematization approaches from existing literature. We use these systematizers to produce concept specs for two concepts---hate-based rhetoric and digital empathy---and evaluate resulting concept specs on content validity and information recoverability.\looseness=-1
\end{abstract}

\section{Introduction}

A critical challenge in evaluating generative AI (GenAI) systems is that many targets of evaluation are broad concepts such as ``reasoning,'' ``fairness,'' or ``creativity.'' Because such concepts admit multiple reasonable definitions across contexts and stakeholder groups, it is often unclear what should be measured~\citep{Wallach2024, Adcock2001}. In practice, evaluations often fail to specify any definition at all~\citep{blodgett2021stereotyping}. Recent work argues that this reflects a missing step in the measurement process: before a broad, \emph{background} concept is operationalized through measurement instruments (e.g., benchmarks, classifiers, or LLM-as-a-judge), it should first be \emph{systematized} into an explicit, structured account of its constituent concepts, the relationships between them, and the observable phenomena that support its measurement~\citep{Jacobs2021, Wallach2024}. A systematized concept can then guide measurement, for example by informing the guidelines given to a judge LLM. And because it is explicit, it can be examined and contested directly: practitioners can ask whether it captures the right notion of the concept, and whether the resulting measurement instrument faithfully reflects it~\citep{Alaa2025medical, Bean2025construct}.\looseness=-1

In practice, however, systematization is cognitively demanding and resource-intensive: it requires making implicit understandings of a concept explicit, enumerating the many ways the concept might manifest, and developing reliable criteria for distinguishing cases that qualify from those that do not. None of this is straightforward. The concept itself may be grasped only intuitively, not in a form that can be readily articulated. The range of relevant cases may be difficult or impossible to map exhaustively. And every definition risks one of two failures: under-specification that leaves scope unsettled, or over-specification that excludes cases it should admit and admits cases it should exclude. To address these challenges, social scientific best practices for systematization recommend synthesizing relevant source material, such as the academic literature~\citep{Goertz2012, corvi-etal-2025-taxonomizing} or the perspectives of diverse stakeholders including domain experts and affected communities~\citep{nguyen2026validating, johnson2026evaluatingaigeneratedimagescultural}. But this is costly work. It is tempting instead to define concepts in terms of what is easy to measure, or to skip explicit definition altogether---asking crowdworkers to annotate examples according to their own implicit understandings and letting a classifier trained on those labels stand in for a systematized concept.\looseness=-1

In this paper, we explore a different response to these challenges: we test whether \textbf{AI-assisted systematization} can make systematization more feasible in practice. Our goal is not to replace human judgment but to provide structured starting points that help practitioners articulate, inspect, and refine what would otherwise remain implicit intuitions or incomplete definitions.\looseness=-1

To enable AI-assisted systematization and assess its quality, we introduce two artifacts grounded in the social scientific literature on systematization~\citep{gerring1999what, hampton2015categories, egre2019concept}. The first is a \emph{concept specification} (or \emph{concept spec}), a structured schema to represent a systematized concept, organized around \emph{patterns} containing \emph{slots} that each take on a set of \emph{slot values}. The second is a \emph{validation worksheet} that articulates desiderata for evaluating concept specs and related artifacts, drawn from social scientific best practice.\looseness=-1

We instantiate AI-assisted systematization in two forms. The first is a \emph{zero-shot systematizer} that prompts an LLM with a description of systematization, the desiderata from our validation worksheet, and the concept spec schema. The second is a \emph{multi-agent systematizer}~(Figure~\ref{fig:system_overview}) that more closely follows manual systematization as described in the literature, decomposing the process into literature review, simulated expert discussion, and iterative specification and validation. This approach draws on multiple perspectives and produces intermediate artifacts that may support human review.\looseness=-1

We evaluate both approaches on two concepts---hate-based rhetoric~\citep{Kennedy2018} and digital empathy~\citep{suh2025sense7}---by comparing the AI-assisted outputs to \emph{reference systematizations} derived from the cited papers and represented in our concept spec schema. We assess them along two dimensions: expert evaluation using our validation worksheet, and \emph{information recoverability}, an approximation of information gain ratio~\citep{Quinlan1986}.\looseness=-1

Our contributions are as follows:
\begin{enumerate}[nosep]
    \item We \textbf{formalize} systematization as the task of transforming a broad background concept into an explicit specification suitable for downstream measurement, introducing the \emph{concept spec} and \emph{validation worksheet}.
    \item We instantiate \textbf{AI-assisted systematization} in two ways: a zero-shot systematizer and a multi-agent systematizer that mirrors manual processes from the literature.\looseness=-1
    \item We empirically \textbf{evaluate} AI-assisted and reference systematizations through two case studies, showing that AI assistance can make systematization more feasible while also revealing its limitations.
\end{enumerate}

\section{Related Work}

\textbf{Measurement, systematization, and validity.}
Our paper builds on recent work locating GenAI evaluation within the social science measurement tradition~\citep{Wallach2024, Adcock2001, Jacobs2021}, in which a \emph{background} concept must be \emph{systematized} before being \emph{operationalized} into measurement instruments. Systematization is what makes it possible to articulate what is being measured, and in turn to assess whether the resulting instrument actually measures it~\citep{Cronbach1955, messick1989validity}. The paper's contribution is to ask whether systematization---the step current GenAI evaluations most often skip---can be made more feasible with AI assistance.\looseness=-1

\textbf{Limits of current evaluation practices.}
A growing literature documents the consequences of this gap in AI evaluation. Benchmarks frequently standardize \emph{how} models are tested without specifying \emph{what} is measured~\citep{raji2021ai, eriksson2025trustaibenchmarks}, and reviews of evaluations find that contested concepts such as ``safety'' and ``robustness'' are often weakly defined---leading evaluations to capture convenient artifacts rather than the intended targets and to prioritize what is easy to test over what matters in deployment~\citep{Alaa2025medical, Bean2025construct, hutchinson2022gaps, agarwal2026fluent}. Recent calls for an ``evaluation science'' of GenAI argue that static benchmarks and ad hoc audits are inadequate for high-stakes, real-world settings~\citep{weidinger2025evaluationscience}. Rather than proposing another benchmark, we focus on the upstream task of turning broad concepts into explicit, auditable specifications.\looseness=-1

\begin{figure*}[!t]
\centering
\begin{mdframed}[innertopmargin=4pt, innerbottommargin=4pt, innerleftmargin=6pt, innerrightmargin=6pt]
\small
\textbf{Pattern 1:} Text that references \texttt{[TARGET\_GROUP]} and conveys \texttt{[HATE\_EXPRESSION]} toward that group.\\
\vspace{-0.7em}

\textbf{Theory:} UN Strategy on Hate Speech 2019, ECRI GPR 15, Kennedy et al. 2022.\\
\vspace{-0.7em}

\textbf{Key Terms:}
\begin{itemize}[itemsep=0.1em, topsep=0.1em, parsep=0pt, partopsep=0pt, leftmargin=1.2em]
    \item \textit{Target Group}: A collection of people identified by at least one protected characteristic.
    \item \textit{Hate Expression}: Language that is derogatory, dehumanising, threatening, or inciting toward a target group.
    \item ...
\end{itemize}

\textbf{Slots:}
\begin{itemize}[itemsep=0.1em, topsep=0.1em, parsep=0pt, partopsep=0pt, leftmargin=1.2em]
    \item \texttt{[TARGET\_GROUP]}:
    \begin{itemize}[nosep, leftmargin=1em]
        \item \textit{Race/Ethnicity}: Groups referenced by race or ethnicity.
        \item \textit{Religion}: Faith communities or religious minorities.
        \item \textit{Gender, Sexual Orientation, Gender Identity}: Women, men, LGBTQ+ or gender-diverse groups.
        \item \textit{Disability, Age, Caste/Class}: Other protected or socially vulnerable groups.
    \end{itemize}

    \item \texttt{[HATE\_EXPRESSION]}:
    \begin{itemize}[nosep, leftmargin=1em]
        \item \textit{Slur}: Recognised derogatory epithet.
        \item \textit{Negative Stereotype}: Claim that a whole group shares an unfavourable trait.
        \item \textit{Dehumanisation/Demonisation}: Comparing a group to animals, disease, or existential threats.
    \end{itemize}
\end{itemize}
\end{mdframed}
\captionsetup{skip=4pt}
\caption{Abridged \textit{concept spec} for hate-based rhetoric from the multi-agent systematizer (full version in App.~\ref{app:hbr-full-concept-specs}).\looseness=-1 }
\label{fig:example_hate_single_pattern}
\vspace{-1em}
\end{figure*}

\textbf{Existing conceptualization and taxonomization work.}
Responsible AI scholars have argued repeatedly that evaluations of social phenomena such as bias~\citep{Savoldi2021, Blodgett2020} and harms~\citep{corvi-etal-2025-taxonomizing, katzman2023taxonomizing, Shelby2023} require explicit conceptualization before they can be measured responsibly~\citep{Jacobs2021, Shahid_2025}. One line of work retrospectively identifies validity problems in existing measurement instruments. Another produces manual systematizations of specific concepts, but this is labor-intensive, which is part of why systematization is often skipped in practice. We ask whether AI assistance can lower this cost, enabling more concepts to be systematized in the first place.\looseness=-1

\section{Problem Formulation}
\label{subsec:concept_spec}

We formulate systematization as the task of producing an explicit representation of a broad background concept. This task has two aspects: the \emph{output}, a representation of the systematized concept, and the \emph{process} of producing it. We formalize the output as a \emph{concept spec} and \emph{auditable process artifacts} as the research reports, discussion logs, or coding manuals that document how the spec was constructed. Both are needed: the spec captures what was systematized, while the process artifacts record the reasoning that produced it.\looseness=-1

A useful concept spec should satisfy three requirements (drawn from our validation worksheet): (1) \emph{clarity}: its components are well-defined; (2) \emph{operationalizability}: the concept is connected to observable phenomena; and (3) \emph{granularity}: it exposes fine-grained distinctions that measurement instruments can target. A prose definition alone cannot meet these requirements; we therefore introduce a structured representation.\looseness=-1

We structure a concept spec as a set of \emph{patterns}, following~\citet{corvi-etal-2025-taxonomizing}, who analyze GenAI outputs in terms of patterns of how utterances relate to the concept of interest (Appendix~\ref{app:concept_spec_grounding}). Let $\mathcal{C}$ denote a background concept, representing a constellation of meanings, interpretations, and intuitions held by stakeholders. A concept spec $\mathcal{S}$ is a set of distinct patterns $\{P_1, P_2, \dots, P_n\}$ that collectively cover the ways $\mathcal{C}$ could manifest as observable phenomena in GenAI outputs. Each pattern $P_i$ describes a specific way the concept manifests, grounded in existing theory and defined over a template $T$, key terms $\mathcal{K}$, slots $\mathcal{Z}$, and theories $\Theta$: $P_i = \langle T, \mathcal{K}, \mathcal{Z}, \Theta \rangle$.\footnote{This schema is not the only valid representation of a systematized concept. Other structures may better suit certain concept types, such as higher-order concepts spanning multiple levels of sub-concepts or concepts defined jointly by user and system actions.} Figure~\ref{fig:example_hate_single_pattern} illustrates a concept spec for hate-based rhetoric; the examples in this section are drawn from this concept.\looseness=-1

\textbf{Pattern Template ($T$).} A descriptive phrase interspersed with \emph{slots} that describes how the concept manifests as observable phenomena. Because each slot can take on different values, the template supports \emph{operationalizability} by capturing multiple ways the concept may manifest without enumerating every linguistic variant.\looseness=-1

\textbf{Key Terms ($\mathcal{K}$).} Important terms used in $T$ and their definitions (e.g., a precise formulation of ``target group''). These anchor the pattern's meaning, supporting \emph{clarity} by reducing ambiguity for stakeholders and downstream measurement instruments.\looseness=-1

\textbf{Slots ($\mathcal{Z}$).} Each slot $Z \in \mathcal{Z}$ maps to a set of possible slot values $\{z_1, z_2, \dots, z_m\}$ or to nested slots. For example, \texttt{[TARGET\_GROUP]} maps to \texttt{race}, \texttt{religion}, etc. Each value comes with a precise definition---containing inclusion and exclusion criteria or examples---so that the concept spec can capture \emph{granular} distinctions.\looseness=-1

\textbf{Theories ($\Theta$).} Existing theories used to justify the pattern's inclusion in $\mathcal{S}$, such as the ``UN Strategy on Hate Speech 2019.'' These references ground the pattern in a recognized account of the concept. They also serve as anchors to the process artifacts: each theory points to specific parts of the research reports, discussion logs, or coding manuals that explain why the pattern was included.\looseness=-1

\begin{figure*}[!t]
    \centering
    \includegraphics[width=0.9\linewidth]{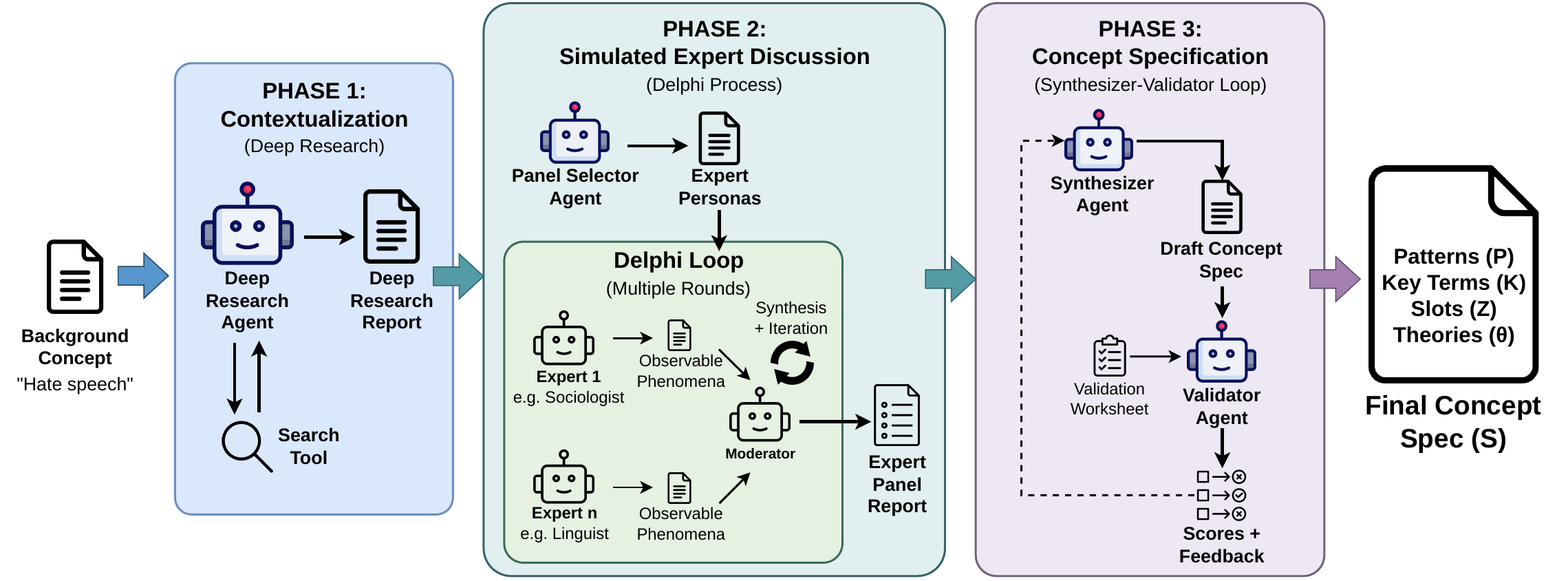}
    \captionsetup{skip=4pt}
    \caption{Overview of multi-agent systematizer: (1) Contextualization: deep research agent synthesizes relevant literature; (2) Simulated Expert Discussion: expert and moderator agents engage in a Delphi process to identify observable phenomena; and (3) Concept Specification: synthesizer and validator agents iteratively refine concept spec.\looseness=-1}
    \vspace{-1em}
    \label{fig:system_overview}
\end{figure*}

\section{AI-Assisted Systematization}

We investigate two approaches to AI-assisted systematization\footnote{We use OpenAI's o3 model for both approaches.} that differ in how explicitly they model the systematization process described in the literature. Studying both lets us ask not only whether AI can help with systematization, but what is gained by modeling the process more explicitly.\looseness=-1

\textbf{Direct Systematizer.}
A reasonable starting point for AI-assisted systematization is to prompt an LLM directly with the information needed to produce and validate a concept spec. Given a background concept, we provide the model with (i) a short description of what systematization entails, (ii) the evaluation criteria from our validation worksheet, and (iii) the concept spec schema from Section~\ref{subsec:concept_spec}. The model produces a concept spec following the schema. The prompt structures the task by specifying both the desired output format and the evaluation criteria, but leaves the process of systematization itself to the model.\looseness=-1

\textbf{Multi-Agent Systematizer.}

The multi-agent systematizer mirrors the structure of manual systematization---reviewing relevant literature, considering multiple perspectives, and iteratively refining a candidate concept spec---through a three-stage architecture (Figure~\ref{fig:system_overview}).

\textit{Phase 1: Contextualization.}
Given a background concept, a deep research agent gathers and synthesizes relevant literature into a detailed report. The report informs all downstream phases.

\textit{Phase 2: Simulated Expert Discussion.}
We simulate a multi-round Delphi-style expert discussion~\citep{dalkey1963experimental}\footnote{A structured communication technique for eliciting and refining expert judgments.}. The aim is to surface a breadth of potentially relevant observable phenomena, rather than to reproduce the views of any particular group. This phase involves three agent roles: a panel selector, simulated experts, and a moderator. The panel selector instantiates a diverse panel of expert personas (e.g., a scholar of marginalization and advocacy), grounded in the deep research report. Simulated expert agents then independently propose observable phenomena, theories, and examples associated with the concept from their ``point of view'' (e.g., the perpetual-foreigner trope, drawing on the microaggressions framework from~\citet{AckermanBarger2020}). The moderator synthesizes these proposals, decides which to merge versus retain as distinct, and feeds the consolidated output into subsequent discussion rounds for iterative refinement. The moderator's synthesis from the last round is passed to the next phase.\looseness=-1

\textit{Phase 3: Concept Specification.}
Finally, a synthesizer agent and validator agent iteratively produce and critique a concept spec, using the deep research report and the synthesized expert outputs as context. In each round, the synthesizer drafts a concept spec and the validator scores it---and intermediate artifacts (research report, expert synthesis)---using the validation worksheet (Appendix~\ref{app:validation_worksheet}). The validator's feedback is incorporated into the next draft, thus iteratively yielding the final concept spec $\mathcal{S}$.\looseness=-1

\section{Evaluation}
Our goal is to evaluate whether AI assistance can support systematization and to characterize the trade-offs across different forms of AI assistance. 
We evaluate three systematization approaches on two case studies---hate-based rhetoric (HBR;~\citealp{Kennedy2018}) and digital empathy (DE;~\citealp{suh2025sense7})---along two dimensions: \textit{content validity} and \textit{information recoverability}. The three approaches are:
\begin{enumerate}[nosep, leftmargin=1.5em]
    \item \textbf{Direct Systematizer}: generates \spec{direct} from a high-level description of the concept (e.g., for HBR: ``Systematize the concept of hate-based rhetoric, where we are concerned with text generated by AI systems.''). It does not produce intermediate artifacts.\looseness=-1
    \item \textbf{Multi-Agent~Systematizer}: uses the same input to generate \spec{multi}, along with intermediate artifacts---the deep research report and simulated expert discussion synthesis---denoted \artifacts{multi}.\looseness=-1
    \item \textbf{Reference Systematizer}: we manually translate each foundational paper's systematization into our concept spec schema, treating the original papers and coding manuals as \artifacts{reference} and producing \spec{reference}. Because the foundational papers present concepts in prose rather than structured form, \spec{reference} should not be read as a gold standard; it reflects how much structured specification can be recovered from existing conceptual work.\looseness=-1
\end{enumerate}

The three concept specs for each case study are provided in App.~\ref{app:hbr-full-concept-specs} and \ref{app:de-full-concept-specs}.\looseness=-1

\subsection{Content Validity via Expert Evaluation} \label{subsec:expert_eval}
We assess content validity through a qualitative evaluation in which two domain experts use a validation worksheet to evaluate both the output (concept spec) and the process (intermediate artifacts).\looseness=-1

\textbf{Expert Evaluators.} 
Both experts have domain expertise relevant to the case studies---hateful content and human-AI relationships---and prior experience with manual systematization and worksheet-based validation. They are included as co-authors of this work in recognition of this contribution but did not participate in generating the systematization artifacts or the downstream analysis. We acknowledge the potential for evaluator bias and treat their judgments as surfacing strengths and weaknesses of different approaches rather than as definitive validation.\looseness=-1

\textbf{Evaluation Worksheet.} Experts assessed each systematization on six attributes. The first three are the same as the concept spec requirements introduced in Section~\ref{subsec:concept_spec}:
\begin{enumerate*}[label=(\roman*)]
\item \textit{clarity},
\item \textit{operationalizability}, and
\item \textit{granularity}.
\end{enumerate*}
The remaining three apply to the systematization process:
\begin{enumerate*}[label=(\roman*), start=4]
\item \textit{provenance}: decision points are documented and traceable, sources are real (not hallucinated);
\item \textit{completeness}: provides confidence in completeness and coverage; and
\item \textit{salience}: relevant to stakeholder needs.
\end{enumerate*}
Experts answered structured questions for each attribute and used their answers to inform a final 1--5 score. Full definitions, rationale, and the worksheet itself are in Appendix~\ref{app:validation_worksheet}.\looseness=-1

\textbf{Procedure.} Each expert was assigned one concept and evaluated all three systematizations (\spec{direct}, \spec{multi}, \spec{reference}). Because the structural and stylistic signatures of the different methods are often recognizable, a strictly double-blind setup is infeasible; we mitigate bias by randomizing presentation order. Evaluation proceeded in two phases. In phase one, experts evaluated the concept spec in isolation, covering the first three attributes. In phase two, they examined the process artifacts, covering the remaining three. Each expert spent approximately 15 hours on the evaluation.\looseness=-1

\subsection{Information Recoverability via Approximate Information Gain Ratio}
We compare concept specs directionally by measuring how much information one concept spec provides about another. Concretely, given a \emph{source} concept spec and a \emph{target} concept spec, we quantify the fraction of the target's uncertainty that is reduced by observing the source. This approximates information gain ratio~\citep{Quinlan1986}, and we call the metric \emph{information recoverability}. High scores indicate that the source concept spec carries substantial information about the distinctions made by the target; low scores indicate that the two concept specs organize the concept space differently.\looseness=-1

\begin{table*}[!t]
\centering
\small
\setlength{\tabcolsep}{5pt}

\begin{minipage}[t]{0.48\linewidth}
\centering
\begin{tabular}{lccc}
\toprule
\multicolumn{4}{c}{\textbf{Hate-Based Rhetoric}} \\
\midrule
\textbf{Attribute} & \textbf{Multi} & \textbf{Direct} & \textbf{Reference} \\
\midrule
\multicolumn{4}{l}{\textit{Evaluated over concept spec}} \\
Clarity & 1 & 1 & 1 \\
Operationaliz. & 2 & 2 & 1 \\
Granularity & 3 & 3 & 1 \\
\midrule
\multicolumn{4}{l}{\textit{Evaluated over process artifacts}} \\
Provenance & 3 & --- & 4 \\
Completeness & 4 & --- & 4 \\
Salience & 4 & --- & 2 \\
\bottomrule
\end{tabular}
\end{minipage}
\hfill
\begin{minipage}[t]{0.48\linewidth}
\centering
\begin{tabular}{lccc}
\toprule
\multicolumn{4}{c}{\textbf{Digital Empathy}} \\
\midrule
\textbf{Attribute} & \textbf{Multi} & \textbf{Direct} & \textbf{Reference} \\
\midrule
\multicolumn{4}{l}{\textit{Evaluated over concept spec}} \\
Clarity & 1 & 4 & 2 \\
Operationaliz. & 3 & 4 & 2 \\
Granularity & 4 & 3 & 1 \\
\midrule
\multicolumn{4}{l}{\textit{Evaluated over process artifacts}} \\
Provenance & 2 & --- & 5 \\
Completeness & 4 & --- & 4 \\
Salience & 4 & --- & 4 \\
\bottomrule
\end{tabular}
\end{minipage}
\captionsetup{skip=4pt}
\caption{Expert evaluation scores (1--5) across case studies, where a higher score is better. Direct produces no auditable process artifacts, so provenance, completeness, and salience are not applicable.\looseness=-1}
\label{tab:expert_eval}
\vspace{-1em}
\end{table*}

\textbf{Intuition.} Evaluating concept specs is challenging because two concept specs may functionally represent the same concept but through a different set of patterns, slots, and slot values.
For example, one concept spec may include a slot \texttt{[Rhetoric]} = \texttt{veiled threat}, while another instead uses \texttt{[Tone]} = \texttt{hostile} and \texttt{[Directness]} = \texttt{implicit}. If the latter's two slot values reliably co-occur with the former's slot value, then we can say the latter concept spec preserves the former's distinctions, albeit through a different structure. By contrast, if one concept spec distinguishes \texttt{reclaimed slur} from \texttt{hostile slur}, but another collapses both into \texttt{profanity}, then the latter cannot faithfully represent the former's distinction.\looseness=-1

\textbf{Procedure.} Given an evaluation dataset (Appendix~\ref{app:dataset_details}), we annotate every sample twice: once using the source spec and once using the target spec. This yields two annotation matrices, $X_{\text{source}}$ and $X_{\text{target}}$, whose rows are samples and columns are slot values from the corresponding spec. Each entry is $1$ if the slot value applies to the sample and $0$ otherwise. We then ask how well the source annotations can recover the target annotations. For each target slot value $g$, let $y_g$ denote the corresponding column of $X_{\text{target}}$, and let $H_g$ denote the Shannon entropy of this vector. We then train a logistic regression model, under stratified $k$-fold cross-validation, to predict $y_g$ from $X_{\text{source}}$. We compute the resulting cross-entropy loss $CE_g$. Slot-level information recoverability  is:
$
\text{Recoverability}_g = 1 - \frac{CE_g}{H_g}.
$\looseness=-1

This quantity approximates information gain ratio with predictive cross-entropy serving as a proxy for conditional entropy.
A score of 1 indicates perfect recoverability; a score of 0 indicates no improvement over predicting the target slot simply from its distribution in the dataset. We report an entropy-weighted average across all target slots $g$ as the overall metric (details in Appendix~\ref{app:functional-adequacy-details}).\looseness=-1

\section{Results: Content Validity}

We analyze the scores and responses produced by experts following the validation worksheet.\looseness=-1

\subsection{Quantitative Results}

Table~\ref{tab:expert_eval} summarizes the expert evaluation scores. Across both case studies, \spec{reference} received lower scores than the AI-assisted systematization concept specs on several attributes. Because both \spec{reference}s were derived from academic papers that were not specifically designed for properties such as operationalizability or granularity, this should not be interpreted as evidence of inferiority. 
Rather, the results are a comparison against a lower-bound faithful representation of what we could recover from available resources.\looseness=-1

\textbf{Hate-based rhetoric.} Both AI-assisted concept specs match each other on the three concept-spec attributes (clarity, operationalizability, granularity), both scoring higher than \spec{reference} on operationalizability and granularity. On the process attributes, \spec{multi} averages slightly higher than \spec{reference} (3.7 vs. 3.3); \spec{direct} cannot be evaluated on these attributes, since it does not produce process artifacts. The two forms of AI assistance thus produce concept specs of comparable quality, but only the multi-agent approach makes the process itself auditable.\looseness=-1

\textbf{Digital empathy.} For DE, \spec{direct} scores notably higher than \spec{multi} on clarity, while the two concept specs are similar over operationalizability and granularity. On the process attributes, \spec{reference} averages higher than \spec{multi} (4.3 vs. 3.3).
These results suggest a trade-off: the direct systematizer can yield a clearer concept spec, while the multi-agent systematizer adds process artifacts at the cost of some concept-spec clarity.\looseness=-1

\subsection{Qualitative Results}

We analyze the experts' qualitative feedback on clarity, operationalizability, and granularity, which are evaluated over the concept spec. Other attribute findings (for the process) are in Appendix~\ref{app:qual-results}.\looseness=-1

\textbf{Inconsistency between a slot and its values reduces \textit{clarity}.}
Across the concept specs produced with AI-assisted systematization, experts noted cases where the slot definition and its enumerated values did not fully align.
For example, in \spec{HBR-multi}, one pattern uses a slot \texttt{[TARGET GROUP]}, which is defined as a ``collection of people identified by protected characteristics such as race, ethnicity, nationality, religion, ...''
This definition was judged to be well-defined; it was defined once for the first pattern and reused consistently in subsequent patterns.
However,
some of these protected characteristics enumerated in the \textit{definition} were omitted as \textit{slot values}, while additional categories such as \texttt{migration status} were included only as slot values.
If there are mismatches between a slot's definition and the values it can take, a downstream operationalization that relies on the slot values would not be able to validly make claims about the slot.
This issue appeared only in the concept specs from AI-assisted systematization, suggesting that it stems from the inconsistency of LLM-based systems.\looseness=-1 %

\textbf{Inconsistency in slot value definitions also reduces \textit{clarity}.}
Although slot value definitions were often quite detailed (see App.~\ref{app:hbr-full-concept-specs} and \ref{app:de-full-concept-specs}), experts noted that
definitions were sometimes inconsistent across a spec. Some definitions left ambiguity (e.g., ``references to groups by race and ethnicity''), some relied on non-comprehensive examples requiring additional inference to arrive at definitions (e.g., trying to infer the boundary between gender and gender identity when the former is defined as ``women, men, non-binary people''
and the latter is defined as ``transgender, non-binary, or gender-diverse groups''),
some were circular (e.g., \texttt{race/ethnicity} defined as ``references to groups by race and ethnicity''), and some were written as regex. Slot value definitions also varied in scope: some were broad (e.g., ``\texttt{religion}: references to faith communities''), while others were narrow (e.g., ``\texttt{gender}: women, men, non-binary people'')---the former creating fuzzy boundaries between slot values, and the latter potentially missing instantiations.
When definitions, examples, and descriptors take such heterogeneous forms within a single spec, slot values become harder to apply consistently.\looseness=-1

\textbf{\textit{Operationalizability} involves trade-offs.} 
While lack of clarity generally impedes operationalizability, experts highlighted two additional nuances. First, while imperfect alignment between slot definitions and their values could nevertheless support downstream measurement, important edge cases would be missed.
For example, in \spec{HBR-multi}, the slot \texttt{[INFERIOR TRAIT]} is defined as an ``alleged inborn or unchangeable negative attribute of a target group,'' with values such as \texttt{low intelligence}, \texttt{moral corruption}, \texttt{criminal nature}, and \texttt{disease/pollution}. These values could support measurement (e.g., as part of annotation guidelines), but instances involving other types of inferiority (e.g., physical or aesthetic) with no slot values would be difficult to classify consistently. 
Second, some definitions lean heavily toward operationalization, but at the expense of conceptual clarity. For instance, regex-style definitions may be easy to apply mechanically but make the underlying concept harder to understand (e.g., ``regex cue \verb+/\b(here for you|listen|understand)\b/+'' for \texttt{emotional support}). 
Such overly narrow operational definitions can lead to brittle downstream usage that fails to capture the broader phenomenon.\looseness=-1

\begin{table*}[t]
\centering
\small
\setlength{\tabcolsep}{5pt}

\begin{minipage}[t]{0.48\linewidth}
\centering
\begin{tabular}{l c c c}
\toprule
\multicolumn{4}{c}{\textbf{Hate-Based Rhetoric}}\\
\midrule
 & \multicolumn{3}{c}{\textbf{Target (Predicted)}} \\
\cmidrule(lr){2-4}
\shortstack{\textbf{Source}\\\textbf{(Predictor)}} & \textbf{Multi} & \textbf{Direct} & \textbf{Reference}  \\
\midrule
\textbf{Multi}           & ---   & 0.505 & 0.358 \\
\textbf{Direct}          & 0.391 & ---   & 0.345 \\
\textbf{Reference}       & 0.372 & 0.432 & ---   \\
\midrule
\end{tabular}
\end{minipage}
\hfill
\begin{minipage}[t]{0.48\linewidth}
\centering
\begin{tabular}{l c c c}
\toprule
\multicolumn{4}{c}{\textbf{Digital Empathy}}\\
\midrule
 & \multicolumn{3}{c}{\textbf{Target (Predicted)}} \\
\cmidrule(lr){2-4}
\shortstack{\textbf{Source}\\\textbf{(Predictor)}} & \textbf{Multi} & \textbf{Direct} & \textbf{Reference} \\
\midrule
\textbf{Multi}         & ---    & 0.277 & 0.172   \\
\textbf{Direct}        & 0.317  & ---   & 0.162   \\
\textbf{Reference}     & 0.138  & 0.124 & ---     \\
\midrule
\end{tabular}
\end{minipage}
\captionsetup{skip=4pt}
\caption{Information recoverability: the fraction of target spec's uncertainty reduced by observing the source spec. Higher scores are better.\looseness=-1}
\label{tab:information_recoverability}
\vspace{-1em}
\end{table*}

\textbf{The concept spec structure enables \textit{granular} analysis, but some specs still lacked sufficient granularity.}
The pattern/slot/value structure supports granular analysis and makes it possible to identify which parts of a concept spec need refinement.
However, some of the concept specs still lacked sufficient granularity.
In \spec{HBR-reference}, overly broad and potentially overloaded definitions led to a lack of granularity (e.g., \texttt{[FRAMING DEVICE]} = \texttt{explicit} defined as ``explicit rhetoric is unambiguous in its potential to be hateful'').
In \spec{DE-direct}, experts noted that the spec was relatively clear (high clarity) but still too coarse (lack of granularity). In \spec{DE-reference}, the use of Likert scale slot values with minimal definitions made it possible to distinguish only between levels of high-level behavior, rather than between more granular types or instances of that behavior.\looseness=-1

\section{Results: Information Recoverability}
Table~\ref{tab:information_recoverability} summarizes the results. Higher scores indicate that the source spec is strongly informative about the target spec. Lower scores do not necessarily indicate lower quality, but that the two specs carve up the concept space differently.\looseness=-1

\textbf{Hate-Based Rhetoric.} Recoverability scores for HBR are moderate overall. When \spec{direct} is the source, it recovers \spec{reference} and \spec{multi} with scores of 0.345 and 0.391. But when \spec{direct} is the \emph{target}, both alternatives have higher recoverability scores---0.432 for \spec{reference} and 0.505 for \spec{multi}. This asymmetry suggests that \spec{direct} has a coarser representation of the concept than the other two concept specs: its distinctions can be better recovered from either of the other concept specs than vice versa. The relationship between \spec{reference} and \spec{multi} is more balanced: \spec{reference} recovers \spec{multi} with a score of 0.372, while \spec{multi} recovers \spec{reference} with a score of 0.358. This indicates these two concept specs share some distinctions but organize others differently.\looseness=-1

\textbf{Digital Empathy.} Recoverability scores for DE are comparatively lower, indicating less shared information across the concept specs. Notably, comparisons involving \spec{reference}---as either source or target---are consistently weak (0.124--0.172), whereas \spec{multi} and \spec{direct} recover each other substantially better (0.277 and 0.317). This pattern suggests that \spec{reference} organizes the concept space differently from the two AI-assisted concept specs, while \spec{direct} and \spec{multi} share a more similar decomposition. Manual inspection supports this: \spec{reference} represents empathy  through Likert-style dimensions, whereas \spec{direct} and \spec{multi} decompose the concept into more granular categorical constituents. %

\section{Discussion}
Several observations follow from our evaluation.\looseness=-1

\textbf{Good systematization is multi-dimensional.} Quality depends on both the concept spec and the process that produced it, and these dimensions can be orthogonal: a process may produce a clear, operationalizable spec while being weakly grounded (e.g., \spec{DE-direct}), while a well-documented process may yield a spec that lacks clarity (e.g., \spec{HBR-reference}). A systematization must be judged on several dimensions at once. In our case studies, the multi-agent systematizer did not uniformly produce better concept specs than the direct systematizer; rather, its main advantage was making the systematization process more auditable.

\textbf{Internal inconsistency is the main failure mode of AI-assisted systematization.} In the AI-assisted concept specs we evaluated, this showed up as misalignment between a slot's intended meaning and its enumerated values, and as heterogeneous formats for slot value definitions. Future AI-assisted systematizers must address internal coherence as well as structural richness.\looseness=-1

\textbf{Recovering systematization from prior conceptual work has its own limits.} Translating a foundational paper into a concept spec is its own task, distinct from both full manual systematization and AI-assisted systematization. It requires interpreting narrative prose, deciding how to organize distinctions into patterns and slots, and handling ambiguities the paper itself leaves implicit. Our reference systematizations sometimes scored lower than the AI-assisted alternatives, which may reflect the difficulty of this work.\looseness=-1

\textbf{AI assistance can make explicit systematization more feasible.} Taken together, these findings shape how we view AI assistance. We do not argue that systematization should be fully automated, nor that the concept spec is the only valid representation. Human judgment remains essential. But in practice, the alternative to AI-assisted systematization is often skipping the step entirely---jumping straight from a background concept to operationalization. The contribution of AI assistance is not a finished product, but a structured starting point for further refinement.\looseness=-1

\textbf{A framework for studying AI-assisted systematization.} Beyond the specific systems we evaluate, this paper contributes a framework for studying AI-assisted systematization broadly. The concept spec provides a common representation across manual and AI-assisted approaches; the validation worksheet provides a way to assess their strengths and weaknesses; and the comparison between direct and multi-agent assistance clarifies the trade-offs of different forms of AI assistance. These are generalizable artifacts for future work on making measurement more explicit, comparable, and auditable.\looseness=-1

\section*{Limitations and Future Work}

Our study has several limitations. First, we study whether AI-assisted systematization can produce concept specs and auditable process artifacts, but we do not study two important downstream questions: whether these specs improve the quality of resulting annotations, and whether real-world practitioners find these approaches more feasible in practice. These questions require separate studies involving downstream measurement instruments and user studies with practitioners.\looseness=-1

Second, we evaluate only two concepts because rigorous expert evaluation of systematized concepts is highly resource-intensive and requires careful engagement. We therefore treat these case studies as an initial test of the framework, and future work is needed to assess generalization across a broader range of concepts. Further, for the same reasons, our evaluation relies on only two expert evaluators, both of whom are co-authors, since this evaluation requires substantial domain expertise that cannot easily be outsourced to non-experts or crowd annotators. Future work should replicate the evaluation with a larger pool of external experts and affected stakeholders.\looseness=-1

Finally, we study only two forms of AI assistance: a lightweight direct systematizer and a more process-faithful multi-agent systematizer. Future work could explore other forms of assistance, including interactive human-in-the-loop workflows. Moreover, the simulated expert discussion in the multi-agent system is not a substitute for real experts or affected communities~\citep{johnson2026evaluatingaigeneratedimagescultural}, and future work should assess which components of the multi-agent architecture are actually necessary, especially since the direct systematizer performs competitively in several cases.\looseness=-1

\section*{Ethics Statement}
This paper discusses high-stakes concepts such as hate speech. Accordingly, the associated concept specs include offensive and contested terms, including slurs, in the Appendix. Full, unredacted, concept specs are included for transparency and to enable robust measurement; they do not reflect the authors' endorsement. AI tools were used to refine the writing and assist with writing code. The authors assume full responsibility for the content and correctness of the paper.

\bibliography{refs}

\appendix

\newpage
\section{Concept Spec Grounding} \label{app:concept_spec_grounding}

We draw on \citet{corvi-etal-2025-taxonomizing}, who leverage speech act theory~\citep{Austin1962} to structure the behaviors of generative AI systems and their impacts. According to this model, each speech act is made up of locutionary, illocutionary, and perlocutionary elements, corresponding respectively to the word choice and ordering, purpose, and real-world impacts of a given utterance. In extending this model for measurement of generative AI system outputs, \citet{corvi-etal-2025-taxonomizing} introduce the notion of illocutionary act \textit{patterns} that, as a set, reflect the different ways the concept of interest manifests in language. Developing the set of patterns enables matching what a given utterance, or system output, is doing as it relates to the concept of interest, without mandating specific lexical or syntactic matching and enumeration of all possible linguistic variants. We leverage this notion of patterns in the concept spec structure.

\section{Example of Simulated Expert  Discussion for Hate-Based Rhetoric}

We illustrate a simulated expert panel discussion generated during the systematization of hate-based rhetoric. Eleven expert personas were selected by the panel selector agent: an international human-rights lawyer, First Amendment scholar, computational linguist, sociolinguist, extremism analyst, content moderator, community advocate, trauma psychologist, AI ethicist, data annotation specialist, and policy regulator. Table~\ref{tab:expert_roles} summarizes the expertise descriptions for a subset of these expert personas. In each round, the simulated experts propose observable phenomena grounded in theory and examples; the moderator then identifies themes, merges overlaps, and synthesizes a consolidated list. An abridged excerpt from the simulated discussion is shown in Figure~\ref{fig:expert-discussion-example}.

\begin{table}[h]
\centering
\scriptsize
\setlength{\tabcolsep}{1pt}
\begin{tabular}{>{\raggedright\arraybackslash}p{0.23\linewidth} 
                >{\raggedright\arraybackslash}p{0.30\linewidth} 
                >{\raggedright\arraybackslash}p{0.45\linewidth}}
\toprule
\textbf{Persona} & \textbf{General Expertise} & \textbf{Sub-Area Expertise} \\
\midrule
Intl.\ human rights lawyer & Global hate speech law & ICCPR, DSA, cross-jurisdiction doctrine \\
First Amendment scholar & U.S.\ free speech law & Civil-liberties limits, overbreadth, platform liability \\
Computational linguist & NLP hate speech detection & Datasets, modeling, evaluation, robustness \\
Sociolinguist & Coded and contextual language & Slurs, dog whistles, multilingual pragmatics \\
Extremism analyst & Online hate ecosystems & Propaganda, symbols, recruitment, evasion \\
\bottomrule
\end{tabular}
\caption{Selected expert personas to systematize hate-based rhetoric.}
\label{tab:expert_roles}
\end{table}

\begin{figure*}[h]
\centering
\begin{tcolorbox}[colback=gray!5!white, colframe=gray!75!black, boxrule=0.5pt]
\small
\textbf{Round 1:}
\begin{itemize}[nosep]
\item \textit{Lawyer:} proposes \textbf{legalistic disenfranchisement}, where hostility is framed as constitutional or security policy rather than insult. Theory: `Abuse of Rights' doctrine in international law (art.~30 UDHR).
\item \textit{Computational linguist:} proposes \textbf{obfuscated hate signals}, including coded slogans, numeronyms, and orthographic variants that preserve hate while evading filters.
\item \textit{Community advocate:} proposes \textbf{identity erasure}, such as systematic misgendering or categorical denial of a group's legitimacy.
\item \textit{Moderator synthesis:} groups these into broader initial families: explicit incitement, coded/euphemistic hate, identity denial, discriminatory policy advocacy, and pseudo-scientific legitimation.
\end{itemize}

\textbf{Round 2:}
\begin{itemize}[nosep]
\item \textit{Extremism analyst:} adds \textbf{demographic-threat narratives} (``replacement'' frames) and \textbf{conspiracy-based vilification} based on \cite{allport1954nature}.
\item \textit{Content moderator:} adds \textbf{moral-panic framing}, especially claims that a target group threatens children or public safety. Drawing on \cite{cohen1972folk}.
\item \textit{Moderator synthesis:} separates ``threat framing'' from ``policy exclusion,'' and distinguishes ``dog whistles/obfuscation'' from broader ``conspiracy narratives.''
\end{itemize}

\textbf{Round 3:}
\begin{itemize}[nosep]
\item Experts converge that hate-based rhetoric includes not only slurs and explicit calls for violence, but also: bureaucratic/legalistic exclusion, pseudo-scientific inferiority claims, coded or obfuscated hate signals, identity erasure and misrecognition, and fear-based mobilization frames like demographic threat and child-protection panic.
\item \textit{Final moderator synthesis:} these are retained as the core observable families for the concept.
\end{itemize}
\end{tcolorbox}
\caption{Abridged excerpt of a simulated expert deliberation during the systematization of hate-based rhetoric. Example utterances are redacted and paraphrased for safety and clarity.}
\label{fig:expert-discussion-example}
\end{figure*}

\section{Information Recoverability: Dataset Details}
\label{app:dataset_details}

To create the sampled dataset from the Gab Hate Corpus and from the SENSE-7 dataset, we took a stratified sample over both datasets, resulting in a sub-sampled dataset of 2000 samples from GHC and a sub-sampled dataset of 764 samples from SENSE-7.

\subsection{Stratified Sampling}
GHC had an overarching annotated category of ``hate,'' and SENSE-7 had an overarching annotated category of ``empathy overall,'' so those were used to determine the strata for stratified sampling.

For GHC, we averaged the hate score (0 or 1) across all annotators for each sample, and then did stratified sampling over the following four buckets of averaged hate values: 0-0.24, 0.25-0.49, 0.50-0.74, 0.75-1.0. We limited the number of samples per bucket to 500.

For SENSE-7, the data is composed of several conversations, where each conversation is multiple user-AI assistant turns.
We treated all the previous turns in a conversation plus an AI assistant turn as a single sample, so each conversation was expanded out into multiple samples.
We performed stratified sampling over three buckets: very poor or poor empathy overall, neutral empathy overall, and good or very good empathy overall.
For this sampling, we were limited by the amount in the smallest bucket to ensure an equal sample size across buckets.

\section{Information Recoverability: Implementation Details}
\label{app:functional-adequacy-details}

This appendix provides additional detail on how the information recoverability metric was computed in practice. For each concept, we begin with a dataset containing examples of occurrences of the concept: Gab Hate Corpus for hate-based rhetoric and SENSE-7 for digital empathy. We then annotate this dataset under each concept spec being compared using an LLM annotator.\looseness=-1

\textbf{LLM annotation procedure.}
For every sample in the dataset, we present the LLM with the full concept spec and instruct it to annotate \emph{every} slot value across all patterns with a binary judgment (1 if the slot value applies to that example, 0 otherwise), together with a brief free-text rationale. For Gab Hate Corpus (HBR), the model annotates the text; for SENSE-7 (DE), it annotates the assistant's final response in the context of the previous turns. The LLM's response is constrained to a structured JSON schema with one field per slot value. This yields an annotation matrix $X$, whose columns correspond to slot values and rows correspond to samples. We use the same LLM (GPT-5.4 with medium thinking) to annotate both source and target concept specs to avoid annotator-specific confounds in our analysis.\looseness=-1

\textbf{Judge validation.}
To validate annotation reliability, we re-annotated both datasets for each concept spec using another strong model (DeepSeek-V4-Pro) and compared these annotations to the GPT-5.4 annotations using pooled Cohen's $\kappa$ over all binary slot judgments. Agreement was moderate overall (mean $\kappa = 0.52$ across the six dataset--systematization pairs, range $0.40$--$0.68$), with higher agreement on Gab Hate Corpus ($\kappa = 0.59$--$0.68$) than on SENSE-7 ($\kappa = 0.40$--$0.48$). This suggests that the annotation is reasonably stable, while also indicating that the digital empathy concept specs are more difficult to annotate consistently.\looseness=-1

\textbf{Predictive modeling setup.} 
For each source--target concept spec pair, we begin with two annotation matrices over the same evaluation dataset: $X_{\text{source}}$, produced by annotating the dataset with the source concept spec, and $X_{\text{target}}$, produced by annotating the dataset with the target concept spec. We then ask whether the source-spec annotations contain enough information to reconstruct the distinctions made by the target spec.

For each target slot value $g$, let $y_g$ denote the corresponding binary annotation vector, i.e., the column of $X_{\text{target}}$ indicating whether $g$ applies to each sample. We train an $\ell_2$-regularized probabilistic logistic regression model that uses $X_{\text{source}}$ to predict $y_g$, and estimate out-of-sample performance using stratified cross-validation. We compute the resulting cross-entropy loss $CE_g$ in bits, and normalize it by the empirical Shannon entropy $H_g$ of the target vector:
\[
\text{Recoverability}_g = 1 - \frac{CE_g}{H_g}.
\]

\textbf{Different slot topologies.}
Not all slots have the same structure. Some slots are \emph{single-choice}: at most one value can be active for a given sample. These are modeled as multiclass prediction problems, with an additional null category when no value is active. Other slots are \emph{multi-label}: multiple values can be active simultaneously. These are modeled as a small bundle of binary prediction problems, one for each slot value, and the corresponding entropy and cross-entropy terms are summed within the slot before normalization. This lets the metric handle both kinds of slots within a single evaluation framework.

\textbf{Additional details.}
All evaluation uses out-of-fold predicted probabilities, so each example is scored only when it is held out from model fitting. We use up to $k=5$ stratified folds, reducing the number of folds automatically when a slot has limited class support. Very rare targets are excluded when there are too few examples to support reliable stratified evaluation. To produce a single concept-level score, we aggregate across target slots using an entropy-weighted average,
\(
1 - \frac{\sum_g CE_g}{\sum_g H_g},
\)
so that slots with greater intrinsic uncertainty contribute proportionally more to the final score.

\section{Qualitative Results for Provenance, Completeness, and Salience} \label{app:qual-results}

\textbf{\textit{Provenance} requires a coherent narrative and is tricky to evaluate over a collection of systematization artifacts.}
Provenance was the one attribute evaluated over systematization artifacts where the artifacts produced from the multi-agent system (a deep research report and a log of the simulated expert discussion and synthesis) consistently scored lower than those produced by the reference methods.
In the worksheet responses, experts justified the score for \spec{multi} by pointing to the fact that the deep research report did not have as much presentation of different theories, cited sources, and academic citations as one might expect. Cited sources were also not necessarily relevant and deep.
In terms of the log of simulated expert discussion and synthesis, there was potential for higher provenance, as both experts noted that the log contained more presentation of theories, and one expert further noted that the concept spec seemed to be well-grounded in the log.
However, as each log was over a hundred pages long, it was infeasible to manually conduct a detailed, rigorous investigation of the entire thing, as the experts noted.
This observation touches on a tension between the fact that more details of a systematization process lead to more opportunities for interrogation, but potentially less ease of comprehension of the process.\looseness=-1

\textbf{\textit{Completeness} and \textit{salience} of systematization artifacts are relatively high.}
All artifacts have completeness and salience scores of 4, except the \artifacts{HBR-reference}, which has a salience score of 2.
For completeness, experts provided general descriptions of comparisons to alternative systematizations and how existing theories were stretched or repurposed.
For salience, surfaced issues include potential stakeholder misunderstandings that could arise due to a lack of clear definitions, poor examples, potential over-stretching of theories for generalizability, and other choices specified in the artifacts.\looseness=-1

\section{Validation Worksheet} \label{app:validation_worksheet}

\subsection{Motivation and Design Rationale}

\subsubsection{What is a good systematized concept?}
We drew on literature from various domains, including political science, linguistics, and policy to understand what constitutes a “good” systematized concept. The ultimate goal was to develop a worksheet/rubric on which to validate systematized concepts developed for the purpose of evaluating generative AI outputs.  

We examined five papers that posit best practices for concept definition, theory development, and application to policy. \cite{gerring1999what}’s “What makes a concept good” and \cite{Sartori1984}’s “Guidelines for Concept Analysis” are seminal works on concept analysis in the political science domain; \cite{hampton2015categories}’s “Categories, prototypes, and exemplars” examines the concept generation process through a psycholinguistic lens. \cite{egre2019concept}’s “Concept utility” takes a practical view and operationalizes concept utility, while \citet{cash2003knowledge} discuss what makes good evidence for policy. These papers reflect a breadth of expertise from multiple domains, selected by members of an interdisciplinary team working on AI evaluation. Despite the differences among the disciplines represented here, the best practices described have much in common. From these, we propose a synthesized set of guidelines for validating a systematized concept. 

\subsubsection{Synthesis}
In the first four papers, \textbf{defining the concept via differentiation} is a core component of creating a concept definition or theory. Differentiation involves identifying the attributes that distinguish a concept from related concepts, ensuring that it is clearly bounded and unique. Sartori describes a generative process for concept definition, which involves collecting definitions, extracting their characteristics, and organizing them meaningfully. This process maps nicely onto Hampton's exemplar process, where specific examples are used to understand the variations and boundaries of a concept. Egré and O'Madagain leverage the use of differentiation in a concept taxonomy to understand the utility of a concept. 

Gerring and Sartori both emphasize the \textbf{careful use of terminology} in concept/theory definition. Gerring highlights the importance of familiarity and resonance, suggesting that terms should be chosen within established usage and should “ring” or “click.” Sartori posits that key terms must be defined, unambiguous, and used consistently.  

Establishing the relationship between the proposed concept definition and neighboring terms is another common theme in the papers. Gerring's field utility involves establishing relationships with neighboring terms within the semantic field. Sartori focuses on relating the term to neighboring words and showing that reconceptualization brings more clarity, not only to the concept of interest, but to adjacent concepts as well. Hampton’s process of categorizing via exemplars is itself a process of understanding a category via its membership and non-membership of objects in related categories.

While Gerring and Sartori each provide insights into the \textbf{utility and consequences} of concept definitions and theories, Egré and O'Madagain offer a formalization for concept utility, balancing the tension between the number of properties in a concept definition (“connotation” in Sartori’s terminology) and the breadth covered by the concept (“denotation”) for the purpose of comparing two different systematizations.

Finally, Cash et al. take a stakeholder-oriented view and discuss three desiderata for evidence-based policy: credibility, legitimacy, and salience. Credibility relates to the scientific rigor of the evidence; in the case of concept systematization, this means applying best practices from concept analysis outlined above. Legitimacy concerns to what extent the evidence considers the varied perspectives of different individuals and groups. 

\subsubsection{Application to validation worksheet}
To create a good systematized concept, several key properties must be considered. These properties ensure that the concept is clear, precise, and useful within its intended context.

\textbf{Clarity, Completeness, and Granularity.} A good systematized concept begins with a clear and concise core definition. Key terms must be defined, used consistently, and be neither vague nor ambiguous. A concept definition must provide sufficient coverage of the background concept within the context of use. A well-differentiated concept can be represented in granular terms that allow analysis at the granular level, as well as the aggregated level. Higher granularity also enables a variety of configurations that can be adapted to stakeholder needs.  

\textbf{Operationalizability, Provenance, and Salience.} Operationalizability (the ability to transform the systematized concept into measurable phenomena) is a key property for stakeholders; even a well-formulated concept can be difficult to apply in practice. Provenance relates to the traceability and justification of decisions made, lending credibility to the systematized concept, while salience ensures that stakeholder perspectives are taken into account. 

\subsection{Full Validation Worksheet}

\includepdf[pages=-,scale=0.9,pagecommand={}]{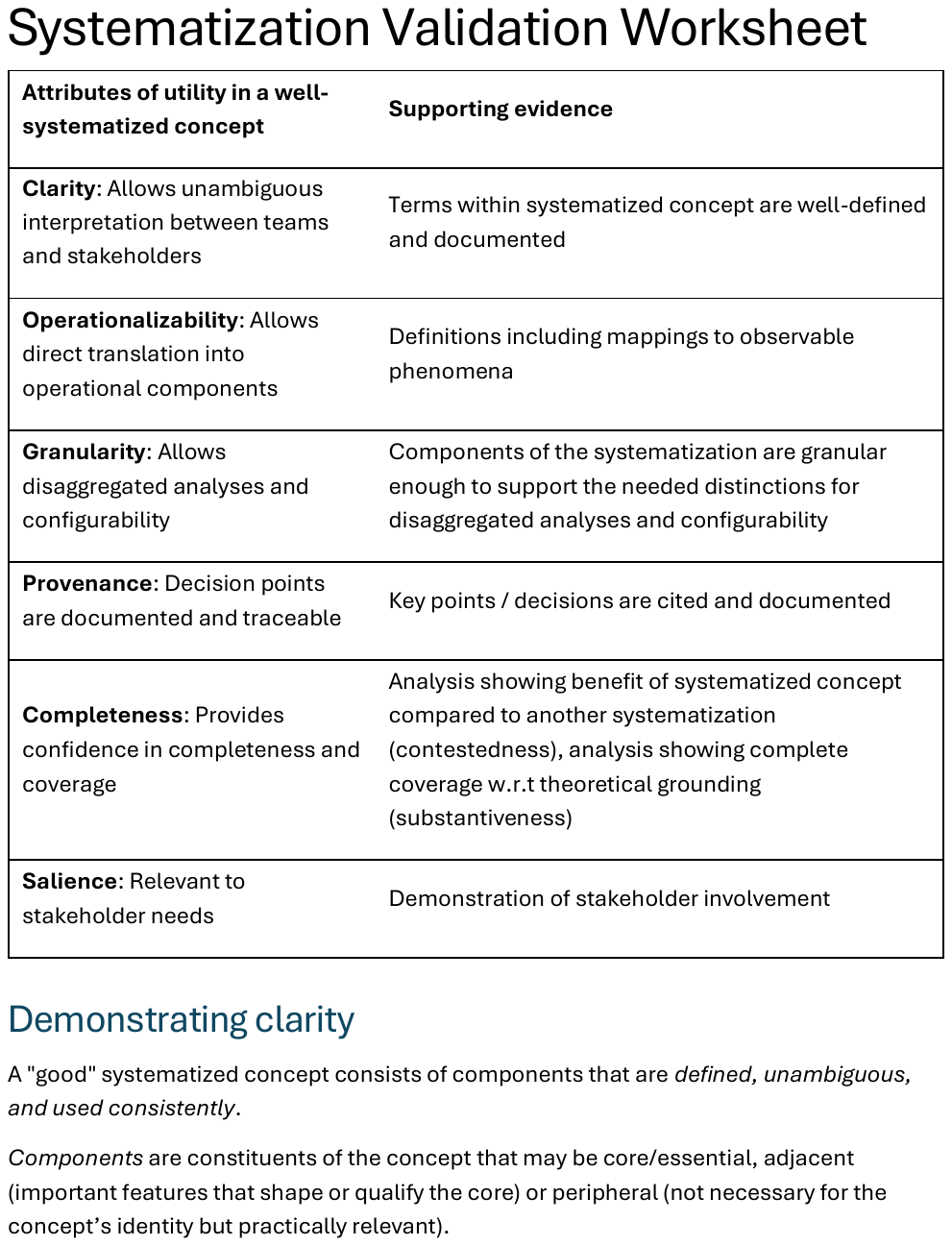}

\section{Concept Specs for Hate-Based Rhetoric} \label{app:hbr-full-concept-specs}

\includepdf[pages=-,scale=0.9,pagecommand={}]{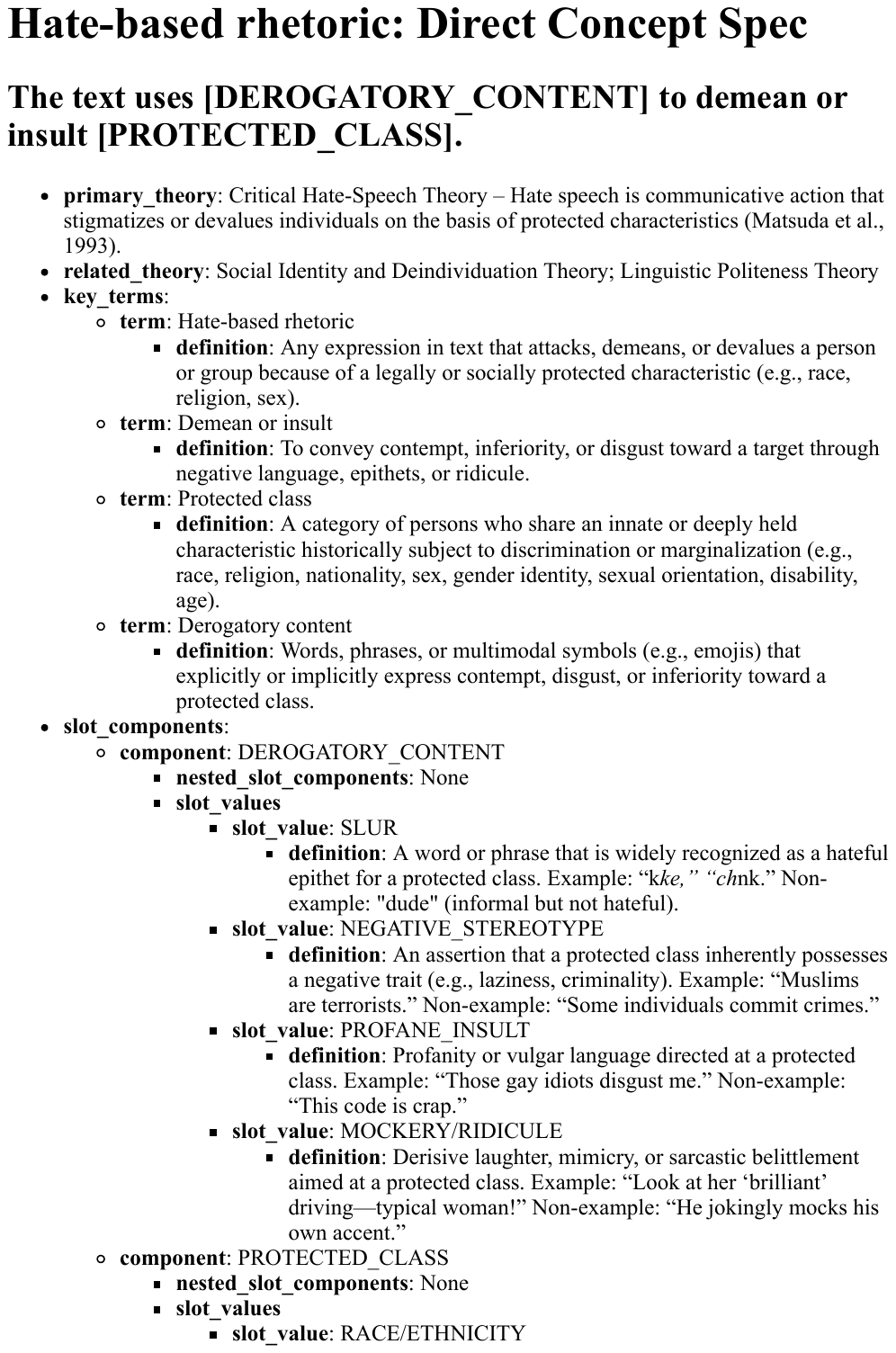}

\section{Concept Specs for Digital Empathy} \label{app:de-full-concept-specs}

\includepdf[pages=-,scale=0.9,pagecommand={}]{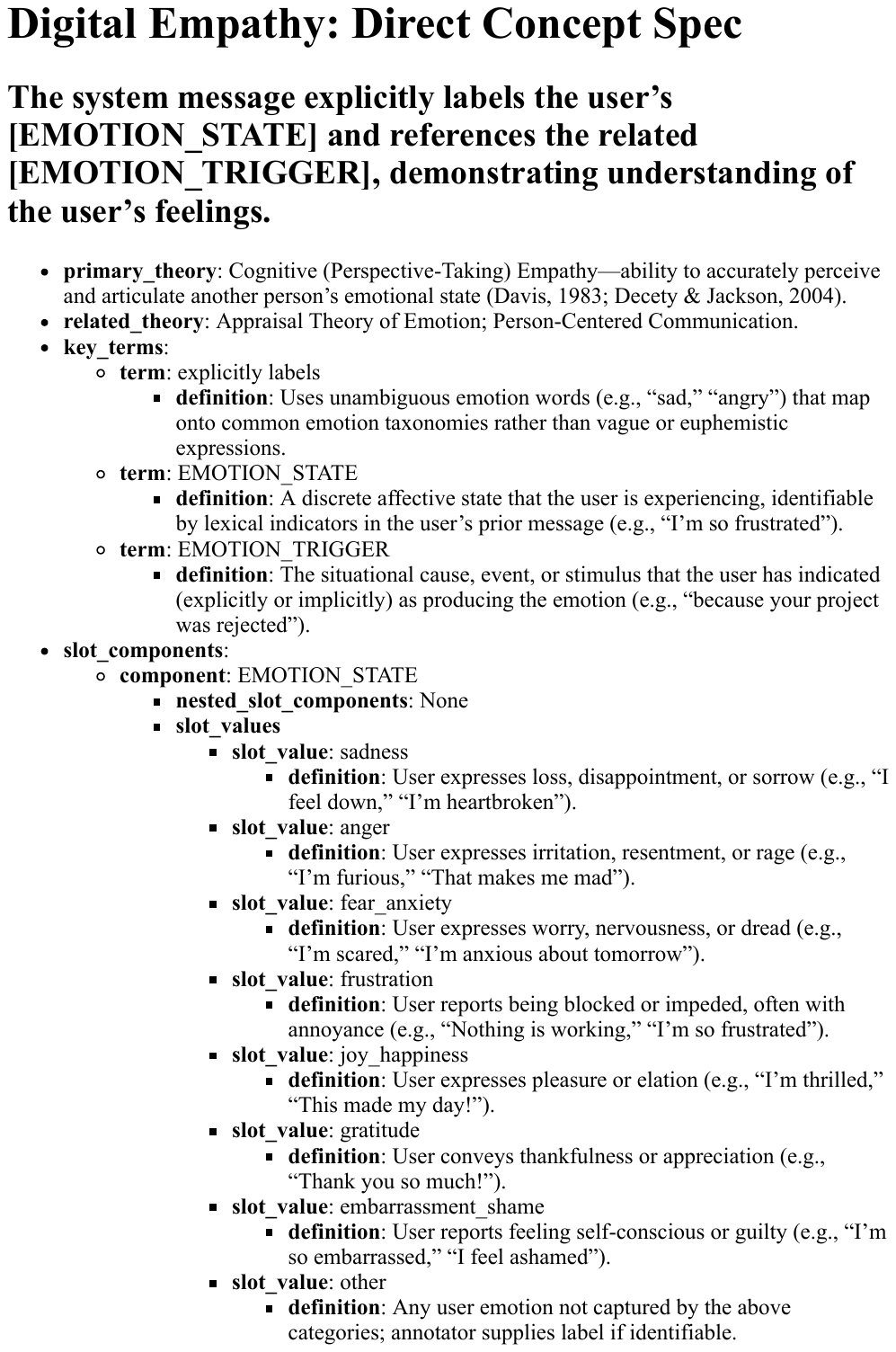}

\end{document}